\documentclass[nohyperref]{article}

\usepackage{microtype}
\usepackage{graphicx}
\usepackage{subfigure}
\usepackage{booktabs} 

\usepackage{hyperref}

\usepackage[accepted]{icml2022}

\usepackage{amsmath}
\usepackage{amssymb}
\usepackage{mathtools}
\usepackage{amsthm}

\usepackage[capitalize,noabbrev]{cleveref}

\theoremstyle{plain}

\theoremstyle{definition}

\theoremstyle{remark}

\usepackage[textsize=tiny]{todonotes}

\usepackage{IEEEtrantools}
\usepackage{balance}
\usepackage{paralist}

\icmltitlerunning{Causal Impact Analysis of the IMF Program on Child Poverty using Causal-Graphical Normalizing Flows}

\begin{document}

\twocolumn[
\icmltitle{Counterfactual Analysis of the Impact of the IMF Program on Child Poverty \\in the Global-South Region using Causal-Graphical Normalizing Flows}



\icmlsetsymbol{equal}{*}

\begin{icmlauthorlist}
\icmlauthor{Sourabh Balgi}{ida}
\icmlauthor{Jose M. Pe{\~n}a}{ida}
\icmlauthor{Adel Daoud}{ias}\\
\icmlauthor{\textnormal{ \small $^1$ Department of Computer and Information Science (IDA), Link{\"o}ping University, Link{\"o}ping, Sweden}}{}
\icmlauthor{\textnormal{ \small $^2$ Institute for Analytical Sociology (IAS), Link{\"o}ping University, Link{\"o}ping, Sweden}}{}\\
\icmlauthor{\textnormal{ \small sourabh.balgi@liu.se}}{ida}
\icmlauthor{\textnormal{ \small jose.m.pena@liu.se}}{ida}
\icmlauthor{\textnormal{ \small adel.daoud@liu.se}}{ias}


\end{icmlauthorlist}



\icmlkeywords{Causal Inference, Structural Causal Models, Normalizing Flows, Counterfactuals, Average Causal Effect (ACE), Conditional Average Causal Effect (CACE), Individual Causal Effect (ICE), Personalised Public Policy Analysis (P$^3$A), Child Poverty}

\vskip 0.3in
]




\begin{abstract}
This work demonstrates the application of a particular branch of causal inference and deep learning models: \emph{causal-Graphical Normalizing Flows (c-GNFs)}. In a recent contribution, scholars showed that normalizing flows carry certain  properties, making them particularly suitable for causal and counterfactual analysis. However, c-GNFs have only been tested in a simulated data setting and no contribution to date have evaluated the application of c-GNFs on large-scale real-world data. Focusing on the \emph{AI for social good}, our study provides a counterfactual analysis of the impact of the International Monetary Fund (IMF) program on child poverty using c-GNFs. The analysis relies on a large-scale real-world observational data: 1,941,734 children under the age of 18, cared for by 567,344 families residing in the 67 countries from the Global-South. While the primary objective of the IMF is to support governments in achieving economic stability, our results find that an IMF program reduces child poverty as a positive side-effect by about 1.2$\pm$0.24 degree (`0' equals no poverty and `7' is maximum poverty). Thus, our article shows how c-GNFs further the use of deep learning and causal inference in AI for social good. It shows how learning algorithms can be used for addressing the untapped potential for a significant social impact through counterfactual inference at population level (ACE), sub-population level (CACE), and individual level (ICE). In contrast to most works that model ACE or CACE but not ICE, c-GNFs enable personalization using \emph{`The First Law of Causal Inference'}.
\end{abstract}

\section{Introduction}\label{sec:intro}

As poverty hinders individuals to flourish, many studies have deepened our understanding of the causes and consequences that push vulnerable groups such as children into poverty \cite{banerjee2011pooreconomics}. However, there is lacking knowledge about how to optimally tailor public policies to alleviate poverty for such vulnerable groups in times of macroeconomic volatility \cite{hallerod2013badgovernance_poorchildren, kino2021adelmlreview}. A key question is then, to what extent can policymakers tailor policies to each child's circumstances and to what extent is it viable to articulate one policy for a population. One of the most powerful international organization today is the International Monetary Fund (IMF), as it is responsible to support macroeconomic stability in the global system. The IMF is part of the United Nations, and hence the mandate of almost all countries in the world. Yet the effects of its programs on children are disputed. Thus, more research is required not only to determine the average affect of IMF on child poverty, but also to develop methods to personalize public policies to children's circumstances~\citep{Daoud2017IMFchildhealth, daoud2018structuraladjustments, DAOUD2019foodagri, akerstrom2019nlp_pe}. 

In this article, we analyze the impact of the IMF program on child poverty at an individual child level by considering the microeconomic factors such as family conditions along with the macroeconomic factors impacted by the austerity of the IMF program. 
This idea of individualization or personalization follows from a policy vision of tailoring interventions to an individual's needs and context. For example,~\citet{banerjee2011pooreconomics} states, 
\begin{quote}
  ``We have to abandon the habit of reducing the poor to cartoon characters and take the time to really understand their lives, in all their complexity and richness.''
\end{quote}
In other words, instead of the conventional Average Causal Effect (ACE) policymaking that ascribes the same treatment to an entire population, researchers need to develop more sensitive frameworks. While articulating policies based on conditional ACE (CACE) is an improvement over ACE policymaking, CACE is still insufficient as it analyzes groups of individuals and hence ignores the individual complexity within-groups~\cite{kunzel2019metalearners}. Thus, CACE still suggests a sub-optimal `one-size-fits-all' policy, and therefore, it is critical to move to the finest level of granularity and perform an Individual Causal Effect (ICE) analysis. Such personalized analysis relies on the tools of causal inference and, more specifically, counterfactual inference~\citep{pearl2009abductionactionprediction,pearl2009causality}. 
Currently, to combat poverty, government officials use mainly an individual's income and similar characteristics to identify vulnerable population and then assess if they are eligible for social welfare programs. Then, a simple rule is often applied: if an individual is eligible, that person will likely receive a fixed `one-size-fits-all' public policy; if the individual is ineligible, no policy is ascribed. Although such public-policy making is highly transparent as it applies what is best on average for a population (i.e., ACE) or group (i.e., CACE) it lacks an adaptability to that individual's needs that is necessary to efficiently combat poverty, ill-health, and other social ills~\citep{DAOUD2018scarcity_abundance_sufficiency}. For effective combating, government officials and others require methods that are able to personalize these policies~\citep{kino2021adelmlreview}. That is, personalized public policy analysis (P$^3$A) requires methods that can move beyond ACE estimation, and into counterfactual inference.

Although there exists a range of ACE and CACE estimation methods powered by machine-learning algorithms \cite{kunzel2019metalearners}, there is a lack of similar ICE methods. 
If the statistical (functional) form among the outcome, treatment, and covariates are known, one could in principle estimate ICE parametrically \cite{pearl2009abductionactionprediction}.
However, in observational studies, functional forms are rarely known due to the complexity of reality and, thus, causal estimation would benefit if machine-learning methods can learn these functions in a data-driven way. To learn these functional forms, recently \citet{anonymous2021cGNF} developed an ICE estimation method, called causal-Graphical Normalizing Flows (c-GNFs)~. This method is a flow-based deep-learning model, that encapsulates the underlying SCM. In other words, c-GNFs enable a scholar to stipulate the causal connections among variables in an SCM, but without making any functional form assumption. The c-GNF is able to approximate these functional forms efficiently in a data-driven manner. Most importantly, because c-GNFs are invertible (they can propel forward and backward through a SCM) and thus calculate noise variables for each individual, it has been shown that c-GNFs enable ICE estimation using \emph{`The First Law of Causal Inference'}~\cite{pearl2009abductionactionprediction}.

The existing development of c-GNFs relies purely on statistical proofs and simulation studies.
In~\citet{anonymous2021cGNF}, scholars developed c-GNFs and benchmarked them against well established methods such as Inverse Probability Weighting~\citep{Rosenbaum1983propensityscoreipw, hernan2009ipw}, Regression-With-Residuals~\citep{wodtke2020rwr}, and Meta-Learners~\citep{kunzel2019metalearners} under controlled simulated experimental settings where the true causal effect is known.
Thus, a c-GNFs real-world applicability for P$^3$A remains uncertain.

In this article we evaluate c-GNFs in a large-scale real-world observational study. Based on substantive assumptions encoded in an SCM, we encapsulate the causal mechanism identifying the impact of the IMF program on child poverty. The data used are observational and thus carries the usual caveats for observational studies. Our study shows how c-GNFs are likely to provide insightful conclusions that are hidden in such large scale data and that are significant for P$^3$A specifically and AI for social good, generally.

Our article supplies at least three contributions. First, it shows how a synthesis of causal inference and deep-neural networks, through c-GNF can further AI for Social Good. This contribution focuses on tailoring macroeconomic policies, minding children's health and living conditions. Our analysis shows how this tailoring can be done empirically and with different personalized-treatment strategies at different population-level personalization granularity: ACE for the planetary (Global-South) level, CACE for the country level, and ICE for individual (child) level.
 
Second, in contrast to previous work~\citep{daoud2019IMFCPeffectheterogeneity}, our findings reveal that the IMF program has beneficial average (ACE) effect. Our CACE finding indicates that the program is most beneficial for India and Bangladesh, and least beneficial for Iraq, Pakistan, Rwanda, and Zambia. 
Third, as virtually all previous IMF studies focus on ACE or CACE for evaluating the effect of macroeconomic policies \cite{Daoud2017IMFchildhealth, daoud2018structuraladjustments}, our research breaks new ground by being the first in evaluating ICE on children. Our individual child level analysis finds that IMF program is beneficial to 61.22$\pm$4.09\% and harmful to 7.25$\pm$4.89\% of a sample of 1,941,734 children from the Global-South population.

Although we apply c-GNFs to the setting of IMF and children, c-GNFs generalize to other settings and P$^3$A. There is a large application space in AI for social good, covering many social challenges in criminal justice, public health, education, and social welfare \cite{lakkarajuMachineLearningFramework2015, potashPredictiveModelingPublic2015, sankaranApplyingMachineLearning2017, ghaniDataScienceSocial2018,  yeUsingMachineLearning2019, daoud2020statisticalmodeling, shibaHeterogeneityCognitiveDisability2021,  shibaEstimatingImpactSustained2021, kino2021adelmlreview}. Thus, one important future task is to evaluate the strengths and weaknesses of c-GNFs, and continue refining their capacities.

\section{Notation, Problem Definition and Assumptions}\label{sec:problemdefinition,assumptions,notations}
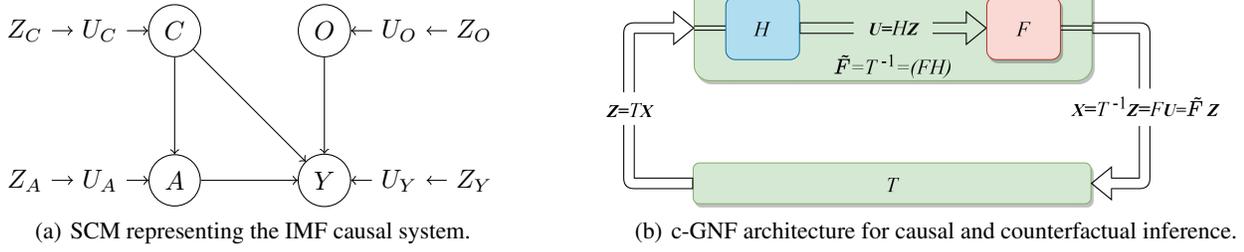
\begin{figure*}[t!]
 \subfigure[SCM representing the IMF causal system.]
 { 
\resizebox{0.4\textwidth}{!}{
\begin{tikzpicture}
\tikzset{vertex/.style = {shape=circle,draw,minimum size=1.5em}}
\tikzset{edge/.style = {->}}
\node[vertex] (C) at (0,-2) {$C$};
\node[vertex] (O) at (2,-2) {$O$};
\node[vertex] (A) at (0,-4) {$A$};
\node[vertex] (Y) at (2,-4) {$Y$};
\node[] (U_C) at (-1,-2) {$U_{C}$};
\node[] (U_O) at (3,-2) {$U_{O}$};
\node[] (U_A) at (-1,-4) {$U_{A}$};
\node[] (U_Y) at (3,-4) {$U_{Y}$};
\node[] (Z_C) at (-2,-2) {$Z_{C}$};
\node[] (Z_O) at (4,-2) {$Z_{O}$};
\node[] (Z_A) at (-2,-4) {$Z_{A}$};
\node[] (Z_Y) at (4,-4) {$Z_{Y}$};
\draw[edge] (Z_C) to (U_C);%
\draw[edge] (Z_O) to (U_O);%
\draw[edge] (Z_A) to (U_A);%
\draw[edge] (Z_Y) to (U_Y);%
\draw[edge] (U_C) to (C);%
\draw[edge] (U_O) to (O);%
\draw[edge] (U_A) to (A);%
\draw[edge] (U_Y) to (Y);%
\draw[edge] (C) to (A);
\draw[edge] (O) to (Y);
\draw[edge] (C) to (Y);
\draw[edge] (A) to (Y);
\end{tikzpicture} 
}
\label{sfig:scm_IMF}
}
\hspace{.05\linewidth}
\subfigure[c-GNF architecture for causal and counterfactual inference.]{
\includegraphics[width=0.51\linewidth,keepaspectratio]{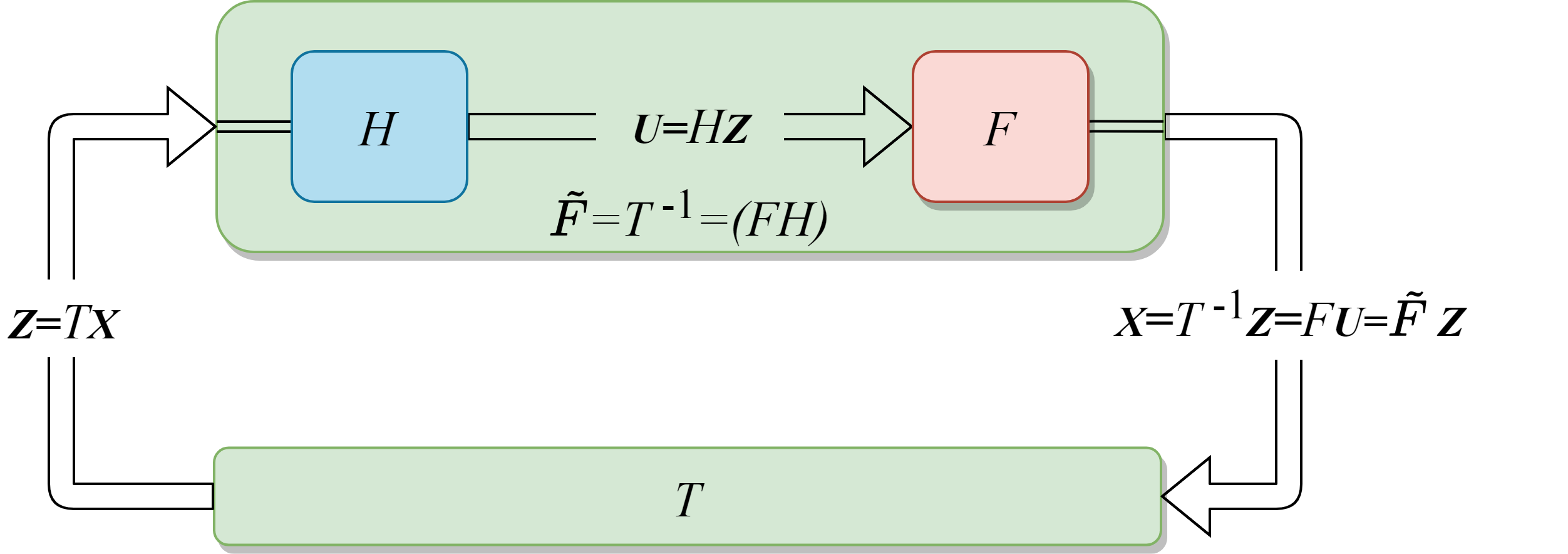}
\label{sfig:c-GNF arch}
}
 \caption{(a) represents the SCM of IMF causal system where $Y$ is the outcome of interest (degree of child poverty), $A$ is the action/intervention/treatment (the IMF program), $C$ are the observed confounders of $A$ and $Y$ (macroeconomic factors), and $O$ are other causes of $Y$ but not of $A$ (microeconomic factors).
For any given observed endogenous variable $X$ in (a), $U_X$ and $Z_X$ respectively denote the unobserved exogenous noises of the true SCM $F{ : }\mathbf{U}{ \rightarrow }\mathbf{X}$ and the \emph{encapsulated-SCM} $\tilde{F}{ = }(FH){ = }T^{-1}{ : }\mathbf{Z}{ \rightarrow }\mathbf{X}$ in Fig.~\ref{sfig:c-GNF arch}, where $H{ : }\mathbf{Z}{ \rightarrow }\mathbf{U}$ denotes an auxiliary transformation, $\mathbf{X}{ = }[O, C, A, Y]^{\text{T}}$, $\mathbf{U}{ = }[U_O, U_C, U_A, U_Y]^{\text{T}}$ and $\mathbf{Z}{ = }[Z_O, Z_C, Z_A, Z_Y]^{\text{T}}$ are the respective vectorial notations. Since $\tilde{F}$ (green) encapsulates $F$ (red) and $H$ (blue) SCMs, i.e., $\tilde{F}{ = }(FH){ = }T^{-1}$, we refer to $\tilde{F}$ as the \emph{encapsulated-SCM}.
The c-GNF models $T{ : }\mathbf{X}{ \rightarrow }\mathbf{Z}$ and readily provides the \emph{encapsulated-SCM} $\tilde{F}{ : }\mathbf{Z}{ \rightarrow }\mathbf{X}$, as $T$ is invertible by construction, facilitating counterfactual inference using \emph{`The First Law of Causal Inference'}~\citep{pearl2009abductionactionprediction, pearl2009causality}.
}
 \label{fig:SimpleCyclicExample}
\end{figure*}

In this section, we discuss the formulation of the causal problem focusing on the relationship between the treatment (IMF program) and the outcome (child poverty). We define our notations and discuss key assumptions.
As our analysis relies on observational data, we are making conditional independence assumptions: the treatment assignment and the outcome are independent conditional on a set of confounders. These confounders have to block all back-door paths \cite{pearl2009causality}. To enforce conditional independence, we rely on existing social-science studies on IMF and child poverty \cite{Daoud2017IMFchildhealth, daoud2018structuraladjustments, daoud2019IMFCPeffectheterogeneity}. Fig.~\ref{sfig:scm_IMF} depicts a conceptual causal-Directed Acyclic Graph (c-DAG) identifying the impact of the IMF program on the degree of child poverty derived from~\citet{daoud2019IMFCPeffectheterogeneity}. 
In Fig.~\ref{sfig:scm_IMF}, $A$ denotes the IMF program as the treatment (exposure, action or intervention) and $Y$ denotes the degree of child poverty as the outcome of causal interest. Moreover, $C$ denotes the observed confounders of $A$ and $Y$, and $O$ are other causes of $Y$ but not of $A$. Specifically, $C$ includes macroeconomic factors such as country's economy, polity, political will, public spending, trade, and governance. These macroeconomic factors are the main confounders. $O$ includes microeconomic and family related factors such as child's age, sex and family wealth. Although these microeconomic variables are unnecessary for adjusting, they are critical to individualize the treatment. 
The c-DAG used in this work involves 32 random variables and it is presented in Fig.~\ref{sfig:dagitty_cdag} of Appendix~\ref{apdx:cDAG}. We note that the c-DAG is constructed with care with the help of social scientists who are the domain experts well aware of the socio-economic literature. However careful the c-DAG is considered, there might be some confounding variables which might be missing and hence, in the assumptions below, we explicitly state the unconfoundedness assumption. Unfortunately, unconfoundedness is an untestable assumption~\cite{Rubin1990unconfoundedness, robins2008ipw} and identifiability of the causal effect is only achievable under unconfoundedness assumption~\cite{pearl2012docalculus}.

\noindent Our main objective in this work is to estimate the causal effect of the treatment (IMF program) at any given population level and personalize treatments depending on the causal effects at that population level or personalization granularity. 
Depending on the personalization granularity, the instantiation of the causal effect might vary, i.e., 
ACE defines the causal effect when personalization granularity is the entire population level (Global-South), CACE defines the causal effect when personalization granularity is a sub-population level (e.g., country, age, gender, etc.), and ICE defines the causal effect when personalization granularity is the individual child level.
Therefore, in the personalization granularity spectrum, we have ACE that offers no personalization and ICE that offers highest personalization at the extremes, with CACE offering intermediate personalization. Formally, ACE, CACE and ICE are defined as follows.
\begin{IEEEeqnarray}{rCl}
     \mathbf{ACE} &= & \mathbf{E}[(Y_{1}-Y_{0})]\IEEEyesnumber \label{eq:ate_tsi_tsj}\enspace ,
\\
\mathbf{CACE}(C{ = }c)
& = & \mathbf{E}[(Y_{1}-Y_{0})|C{ = }c]\IEEEyesnumber\label{eq:cate_tsa_tsb}\enspace ,
\\
 \mathbf{ICE}_i(C{ = }c, O{ = }o) &=& (Y_{1}-Y_{0})|C{ = }c, O{ = }o\IEEEyesnumber\label{eq:ite_tsa_tsb}\enspace ,\enspace 
\\
 \mathbf{ICE}_{it}(C{ = }c, O{ = }o) &=& (1_{(Y_{1}>1)}-1_{(Y_{0}>1)})|C{ = }c, O{ = }o\enspace,\IEEEnonumber\\
 \IEEEyesnumber\label{eq:ite_i_tsa_tsb}
\end{IEEEeqnarray}
where $Y_{a}$ denotes the potential outcome of $Y$ under the intervention $A{ = }a$. Eqs.~\eqref{eq:ite_tsa_tsb} and \eqref{eq:ite_i_tsa_tsb} differ in that the former considers the actual degree of child poverty, whereas the latter considers the threshold $>1$ to obtain an indicator of child poverty, i.e., degrees of 2 to 7 indicate poor and degrees of 0 and 1 indicate not poor, a simplifying assumption considered in~\citet{daoud2019IMFCPeffectheterogeneity}. 

For any given personalization granularity, the treatment assignment condition identifies the optimal treatment to prescribe at the given personalization granularity. In our analysis, we consider six treatment strategies at three personalization granularities as summarized in Table~\ref{table:imf_ts_6} of Appendix~\ref{apdx: ts_summary}, i.e., Global-South (no personalization), country (intermediate personalization), and individual child (finest personalization) levels.  
The six treatment strategies from our study are described as follows:

\begin{inparaenum}[1.]
\qquad\item \textbf{$TS_0$} : The IMF program is discouraged ($A{=}0$) for the entire Global-South. Then, there is no personalization at any granularity.

\qquad\item \textbf{$TS_1$} : The IMF program is encouraged ($A{=}1$) for the entire Global-South. Then, there is no personalization at any granularity.

\qquad\item \textbf{$TS_{Ob}$} : The IMF program is encouraged ($A{=}1$) or discouraged ($A{=}0$) for the entire country based on the observed treatment, i.e., the countries treated get encouraged and the rest do not. 

\qquad\item \textbf{$TS_C$} : The IMF program is encouraged ($A{=}1$) or discouraged ($A{=}0$) or neither for the entire country based on the country's CACE in Eq.~\eqref{eq:cate_tsa_tsb}. Then, there is personalization at country level.

\qquad\item \textbf{$TS_{I}$} : The IMF program is encouraged ($A{=}1$) or discouraged ($A{=}0$) or neither based on the child's ICE in Eq.~\eqref{eq:ite_tsa_tsb}. Then, there is personalization at child level.

\qquad\item \textbf{$TS_{It}$} : The IMF program is encouraged ($A{=}1$) or discouraged ($A{=}0$) or neither based on the child's ICE in Eq.~\eqref{eq:ite_i_tsa_tsb}. Then, there is personalization at child level.
\end{inparaenum}

We note that the expectations in Eqs.~\eqref{eq:ate_tsi_tsj} and~\eqref{eq:cate_tsa_tsb} are with respect to the samples from the interventional distributions $P(Y_{a})$. 
However, the fundamental problem of causal inference dictates that the counterfactual outcomes under the unobserved interventions are always missing, i.e., the samples from $P(Y_{a})$ are not readily available to compute the expectations in  Eqs.~\eqref{eq:ate_tsi_tsj} and~\eqref{eq:cate_tsa_tsb}. These missing values can be computed using the observational data under the following six main assumptions:
\begin{inparaenum}[(a)]
\item unconfoundedness or no unobserved confounders~\citep{Rubin1990unconfoundedness}, 
\item positivity or overlap or common support or extrapolation~\citep{Rosenbaum1983propensityscoreipw},
\item conditional ignorability or exchangability~\citep{Rosenbaum1983propensityscoreipw},
\item consistency~\citep{ROBINS1986gcom, cole2009consistency, vanderweele2009concerningconsistency},
\item no interference or Stable Unit-Treatment Value Assumption (SUTVA)~\citep{cox1958planning}, and
\item modularity or independent mechanisms or autonomy or invariance~\citep{pearl2009causality, peters2017elementsofcausalinference}.
\end{inparaenum}
As our study relies on observational data, some of these assumptions, in particular unconfoundedness and positivity, are untestable. This limitation applies to all observational study. Therefore, any observational study results should be evaluated, jointly with the plausibility of these causal identification assumptions. If one believes the assumptions lack plausibility, then it is possible to make improvement to the unconfoundedness and positivity assumptions by changing the c-DAG and collecting more data. Even though these improvements might change the results, the general mechanics of applying c-GNF to P$^3$A specifically and AI for social good, generally, remains the same.

For the c-DAG in Fig.~\ref{sfig:scm_IMF}, the observational data can at best help modeling the observational joint distribution as 
\begin{IEEEeqnarray}{rCl}
P(O, C, A, Y) = P(Y|A, C, O) P(A|C) P(C) P(O) ~.\enspace\hfill\IEEEyesnumber
\label{eq:scm_imf_obs_dist}
\end{IEEEeqnarray}
As the c-DAG considered assumes unconfoundedness, the $do$-calculus~\citep{pearl2012docalculus} provides a way to identify a valid adjustment set and expresses $P(Y_{a})$ from the components of the observational joint distribution in~\eqref{eq:scm_imf_obs_dist} as,
\begin{IEEEeqnarray}{rCl}
P(Y_{a}) &= \sum_{\{c,o\}}
& P(Y|A{ = }a,C{ = }c,O{ = }o) 1_{(A{ = }a)} \IEEEnonumber\\
& & P(C{ = }c) P(O{ = }o) \enspace,\hfill\IEEEyesnumber
\label{eq:scm_imf_int_dist}
\end{IEEEeqnarray}
where the probability term corresponding to the independent mechanism of the intervened variable is replaced by the indicator function $1_{({A}{ = }a)}$ denoting the intervention ${A}{ = }a$ (modularity assumption). 
Even with the identified adjustment set $\{C\}$ and the non-parametric expression for the interventional distribution in  Eq.~\eqref{eq:scm_imf_int_dist}, the computation of ACE or CACE in Eqs.~\eqref{eq:ate_tsi_tsj}-\eqref{eq:cate_tsa_tsb} still requires a complex and potentially time-consuming summation over the adjustment set variable $\{C\}$. 
Even though $do$-calculus identifies a non-parametric expression for ACE or CACE, $do$-calculus still lacks the estimation of ICE.
However, in the next section, we see that it is possible to easily draw the samples from the distributions in Eqs.~\eqref{eq:scm_imf_obs_dist} and~\eqref{eq:scm_imf_int_dist} using causal-Graphical Normalizing Flows (c-GNFs) without the need of the adjustment set or Eq.~\eqref{eq:scm_imf_int_dist}, using just \emph{`The First Law of Causal Inference'}~\citep{pearl2009causality, pearl2018bookofwhy}. Hence easily estimating ACE, CACE and ICE in Eqs.~\eqref{eq:ate_tsi_tsj}-\eqref{eq:ite_i_tsa_tsb} using Monte-Carlo expectation estimation from the samples drawn from the c-GNF.

\section{SCM, encapsulated-SCM and c-GNF}\label{sec:gnf}
Here, we briefly present the c-GNFs and describe their connection to SCMs. An SCM consists of a set of assignment equations describing the causal relations between the random variables of a causal system represented by a causal-Directed Acyclic Graph (c-DAG), such as
\begin{IEEEeqnarray}{rll}
X_i := f_i(PA_i, U_i) \text{ with } i {=} 1,\ldots,d\label{seq:scm}\IEEEyesnumber\enspace ,
\end{IEEEeqnarray}
where $\{X_i\}_{i{ = }1}^d$ represents the set of observed endogenous random variables, $PA_i$ represents the set of variables that are connoting parents of $X_i$, $\{U_i\}_{i{ = }1}^d$ denotes the set of exogenous noise random variables, and $\{f_i\}_{i{ = }1}^d$ denotes the set of functions (independent mechanisms) that generate the the endogenous variable $X_i$ from its observed causes $PA_i$ and noise $U_i$. 
An SCM is also referred as Functional Causal Model (FCM) due to the functional mechanisms $\{f_i\}_{i{ = }1}^d$. An SCM can be seen as a transformation $F{:}\mathbf{U}{ \rightarrow }\mathbf{X}$ such that $\mathbf{X}{ = }F(\mathbf{U})$ where $\mathbf{U}{ = }[U_1,\ldots,U_d]^\text{T}{ \in }\mathbb{R}^d$ and $\mathbf{X}{ = }[X_1,\ldots,X_d]^\text{T}{ \in }\mathbb{R}^d$.
Let $\mathcal{G}$ represent a c-DAG with $\{X_1,\ldots,X_d\}$ as the set of nodes and adjacency matrix $\mathcal{A_G}{ \in }\{0,1\}^{d{ \times }d}$. Let $P_\mathbf{X}(\mathbf{X})$ represent the joint distribution over the endogenous variables $\mathbf{X}$, which factorizes according to $\mathcal{G}$ as
\begin{IEEEeqnarray}{rll}
P_\mathbf{X}(\mathbf{X}) = \Pi_{i{ = }1}^{d}P({X_i}|PA_i)\enspace,\label{seq:bayesfactor}\IEEEyesnumber\IEEEyesnumber
\label{eq:P_x joint distn bn}
\end{IEEEeqnarray}
where $PA_i{ = }\{X_j{ : }\mathcal{A_G}_{i,j}{ = }1\}$ denotes the set of parents of the vertex/node $X_i$ in $\mathcal{G}$. As we see in the next section, the factorized joint distributions in Eqs.~\eqref{eq:scm_imf_obs_dist} and~\eqref{eq:P_x joint distn bn} can be modeled using an autoregressive model parameterized by a deep-neural-network $\theta$, which we denote by $P_\mathbf{X}(\mathbf{X};\theta)$. Such a model is what is called as c-GNF.

\subsection{Causal-Graphical Normalizing Flows (c-GNFs)}
Normalizing Flows (NFs)~\citep{tabak2010nf, tabak2013nf, rezende2015variationalNF, kobyzev2020NF, papamakarios2021NF_pmi} are a family of generative models with tractable distributions where both sampling and
density evaluation can be efficient and exact. 
A NF is a flow-based model with transformation $T{ : }\mathbf{X}{ \rightarrow }\mathbf{Z}$ such that $\mathbf{Z}{ = }T(\mathbf{X})$, where $\mathbf{Z}{ = }[Z_1,\ldots,Z_d]^\text{T}{ \in }\mathbb{R}^d$ represents the base random variable of the flow-model with the base distribution $P_\mathbf{Z}(\mathbf{Z})$ that is usually a $d$-dimensional standard normal distribution for computational convenience and ease of density estimation. Hence the name normalizing flow. The defining properties of $T$ are, (i) $T$ must be invertible with $T^{-1}$ as the inverse generative flow such that $T^{-1}{ : }\mathbf{Z}{ \rightarrow }\mathbf{X}$, and (ii) $T$ and $T^{-1}$ must be differentiable, i.e., $T$ must be a $d$-dimensional diffeomorphism~\citep{milnor1997diffeomorphism}. 
Under these properties, from the change of variables formula, we can express the endogenous joint distribution $P_\mathbf{X}(\mathbf{X})$ in terms of the assumed base distribution $P_\mathbf{Z}(\mathbf{Z})$ as
\begin{IEEEeqnarray}{rll}
P_\mathbf{X}(\mathbf{X}){ = }P_\mathbf{Z}(T\mathbf{X}) |\mathrm{det} J_{T}(\mathbf{X})|\enspace .\hfill\IEEEyesnumber
\label{eq:p_x joint distn nf}
\end{IEEEeqnarray}
Calculating the joint density of $P_\mathbf{X}(\mathbf{X})$ requires the calculation of the determinant of the Jacobian of $T$ with respect to $\mathbf{X}$, i.e., $\mathrm{det} J_{T}(\mathbf{X})$. It is, therefore, advantageous for computational reasons to choose $T$ to have an autoregressive structure such that $J_{T}(\mathbf{X})$ is a lower-triangular matrix and $\mathrm{det} J_{T}(\mathbf{X})$ is just the product of the diagonal elements. 
Autoregressive Flows (AFs)~\citep{kingma2016IAF, Papamakarios2017MAF, Huang2018NAF} are NFs that model an autoregressive structure.
AFs are composed of two components, the transformer and the conditioner~\citep{papamakarios2021NF_pmi}. Under the assumption of a strictly monotonic transformer, AFs are \emph{universal density estimators}~\citep{Huang2018NAF}. 
Among these AFs, Graphical Normalizing Flows (GNFs)~\citep{wehenkel2020GNF} facilitate the use of a desired DAG representation as opposed to an arbitrary autoregressive structure. 
For causal inference, it is crucial that the DAG has a causal interpretation.
~In this work, we assume the true c-DAG $\mathcal{G}$ for the GNFs, hence the name causal-GNFs (c-GNFs). 
In c-GNFs, we use Unconstrained Monotonic Neural Network (UMNN)~\citep{wehenkel2019UMNN}, a strictly monotonic integration based transformer along with $\mathcal{G}$ for the graphical conditioner to ensure c-GNF is an \emph{universal density estimators}~\citep{Huang2018NAF}.

As Markov equivalent DAGs induce equivalent factorizations of the observational joint distributions, GNFs that use Markov equivalent DAGs represent the same observational joint distribution. This is problematic for causal and counterfactual inference, as different GNFs may represent different interventional joint distributions. In other words, to perform exact causal inference as we do in this work, the GNF needs to use casual-DAG $\mathcal{G}$. 
Note that only a GNF with the true c-DAG $\mathcal{G}$ encapsulates the true SCM, thus satisfying the modularity assumption, which is necessary for correct causal and counterfactual inference. This is formalized as
\begin{IEEEeqnarray}{rll}
\mathbf{X}{ = }F(\mathbf{U}){ = }F(H(\mathbf{Z})){ = }F{ \cdot }H(\mathbf{Z}){ = }\tilde{F}(\mathbf{Z}){ = }T^{-1}(\mathbf{Z})\enspace \enspace\IEEEyesnumber
\label{eq:scm_cgnf}
\end{IEEEeqnarray}
where $H{ : }\mathbf{Z}{ \rightarrow }\mathbf{U}$ is an auxiliary transformation such that $U_i{ = }h_i({Z_i})$ for $i{ = }1,\ldots,d$, i.e., $\mathbf{U}{ = }H(\mathbf{Z})$ without loss of generality. 
It follows from Eq.~\eqref{eq:scm_cgnf} that the c-GNF $T^{-1}{ = }\tilde{F}{ = }F{ \cdot }H$ encapsulates the true SCM $F$ as $T$ and $T^{-1}$ both encode $\mathcal{G}$ in the graphical conditioner, thus providing a way to indirectly model $F$ without making assumptions on $\mathbf{U}$ or the functional causal mechanisms $F$ or the auxiliary transformation $H$. This sets apart the c-GNFs from most other models for causal inference.

Training a c-GNF amounts to training the deep neural networks that parameterize the transformers and conditioners. This is typically done by maximizing the $\log$-likelihood of the training dataset $\{\mathbf{X}^\ell\}^{N_{train}}_{\ell{ = }1}$, which is expressed as shown below, by using Eq.~\eqref{eq:p_x joint distn nf} for the summation term $P_\mathbf{X}(\mathbf{X}^\ell;\theta)$
\begin{IEEEeqnarray}{rll}
\mathcal{L(\theta)}{ = }&\sum_{\ell{ = }1}^{N_{train}}\mathrm{log}P_\mathbf{X}(\mathbf{X}^\ell;\theta)\IEEEyesnumber\enspace , 
\label{eq:cgnf mll fn}
\end{IEEEeqnarray}
where $\theta$ denotes the parameters of the deep neural networks of the UMNN transformer and graphical conditioner, optimized using stochastic gradient descent.
We utilize the Gaussian dequantization trick from~\citet{anonymous2021cGNF} presented in Algorithm~\ref{alg:dequatization} of Appendix~\ref{apdx:GDeq} to model discrete variables into NFs using the fact that NFs are strongest in modeling Gaussian distributions seamlessly. 
Fig.~\ref{sfig:c-GNF arch} shows the c-GNF architecture for causal and counterfactual inference using $T$ and $T^{-1}$.

\subsection{Counterfactual Inference and \emph{`The First Law of Causal Inference'}}
Since a c-GNF encapsulates the true SCM, this enables us to compute the potential outcomes under all the treatment strategies described in Section~\ref{sec:problemdefinition,assumptions,notations} using \emph{`The First Law of Causal Inference'}~\citep{pearl2009abductionactionprediction, pearl2009causality, pearl2018bookofwhy}. This law essentially provides a framework to identify the unit level potential outcomes using an SCM, thereby addressing the fundamental missing value problem of counterfactual inference. The law states that the missing/unseen potential outcome $Y$ for a unit $\mathbf{Z^\ell}$ under the treatment strategy $TS_a$, denoted as $Y_{TS_a}(\mathbf{Z}^\ell)$, can be computed as
\begin{IEEEeqnarray}{rll}
Y_{TS_a}(\mathbf{Z}^\ell){ = }Y_{\tilde{F}_{TS_a}}(\mathbf{Z}^\ell)\enspace,\IEEEyesnumber\label{eq:first law of causal inference}
\end{IEEEeqnarray}
where $Y_{\tilde{F}_{TS_a}}(\mathbf{Z}^\ell)$ represents the actual outcome $Y$ observed from the mutilated-SCM $\tilde{F}_{TS_a}$ that is obtained by removing the causes of $A$ from the \emph{encapsulated-SCM} $\tilde{F}$, and setting $A$ to the value dictated by $TS_a$. Essentially, this law involves the following three steps.
\begin{inparaenum}

    \item \textbf{Abduction}: For a given observed unit-individual evidence $\mathbf{X}^\ell$, the respective exogenous noise $\mathbf{Z}^\ell$ corresponding to $\mathbf{X}^\ell$ is recovered from the encapsulated-SCM $\tilde{F}$. In the case of a c-GNF, we have $\mathbf{Z}^\ell{ = }T\mathbf{X}^\ell{ = }\tilde{F}^{-1}\mathbf{X}^\ell$ where the \emph{encapsulated-SCM} noise $\mathbf{Z}^\ell$ represents the unobserved yet unique identifier of an individual (e.g., DNA) of the $\ell^{th}$ individual with observed evidence $\mathbf{X}^\ell$.
    
    \item \textbf{Action}: The action or intervention on $A$ dictated by the treatment strategy $TS_a$ is performed after mutilating the structural equation corresponding to ${A}$ in the encapsulated-SCM $\tilde{F}$ (from the modularity assumption), which results in the mutilated-SCM $\tilde{F}_{TS_a}$. 
    
    \item \textbf{Prediction}: The recovered noises $\mathbf{Z}^\ell$ from abduction are re-propagated through $\tilde{F}_{TS_a}$ and the outcomes $Y_{\tilde{F}_{TS_a}}(\mathbf{Z}^\ell)$ observed are nothing but the potential outcomes $Y_{TS_a}(\mathbf{Z}^\ell)$ we were interested to begin with.
\end{inparaenum}    

\section{Experimental Results and Discussion}\label{sec:experiments}
\begin{figure*}[ht!]
\includegraphics[width= \textwidth,height=\linewidth,keepaspectratio]{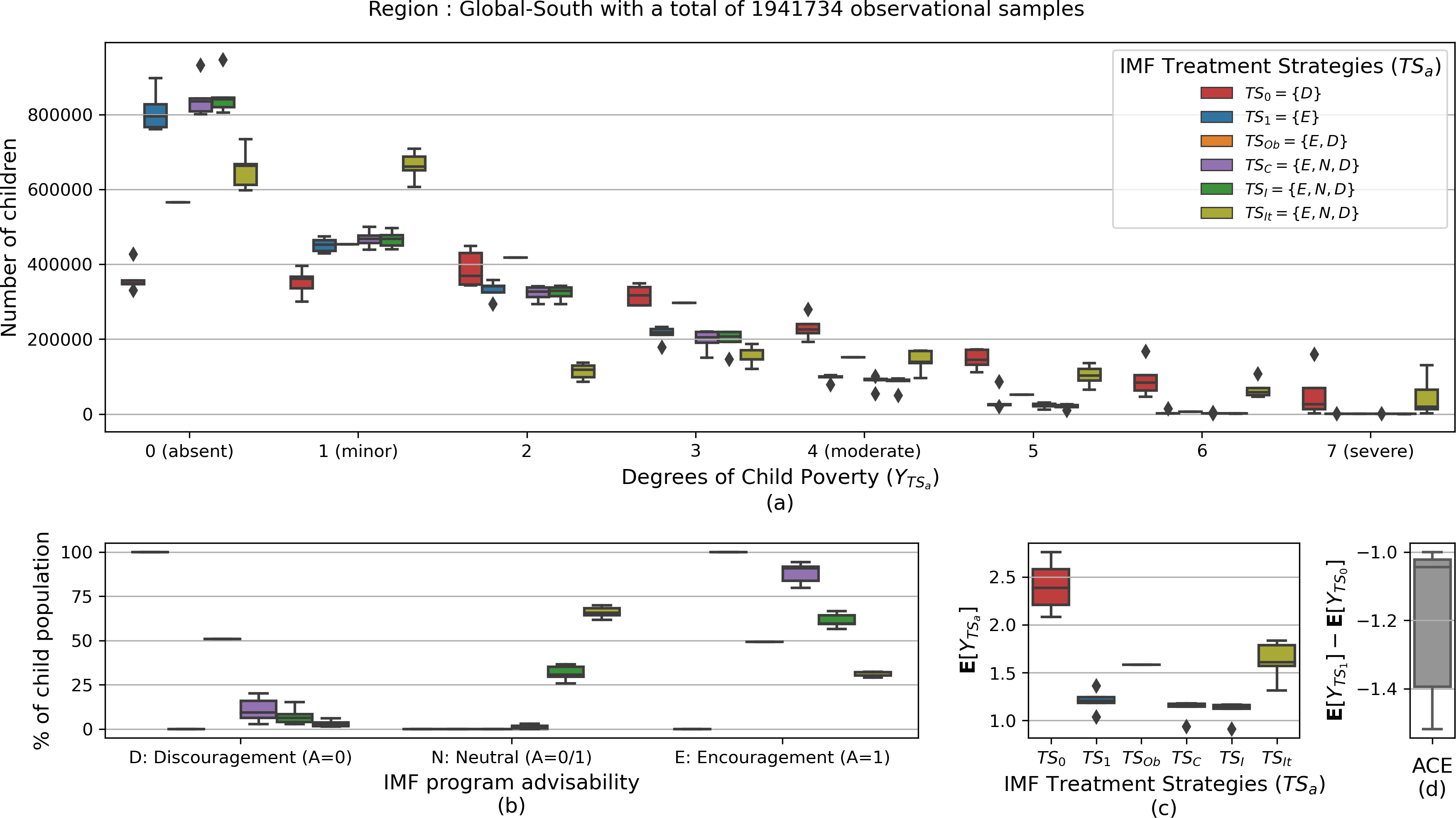}
\caption{All the boxplot results are reported from 5 different random seeded simulations to indicate the variability in the c-GNF results across multiple simulations. (a) presents the number of samples in each of the 8 possible degrees of child poverty across different treatment strategies in the entire Global-South. 
(b) indicates the advisability of the IMF program for each treatment strategy in the entire Global-South. 
(c) indicates the average degree of child poverty for each treatment strategy in the entire Global-South. 
(d) indicates the ACE estimated in the entire Global-South. 
}\label{sfig:IMF_CP_d32_000_GlobalSouth_deg_sim_d}
\end{figure*}
\begin{figure*}[ht!]
\includegraphics[width= \textwidth,keepaspectratio]{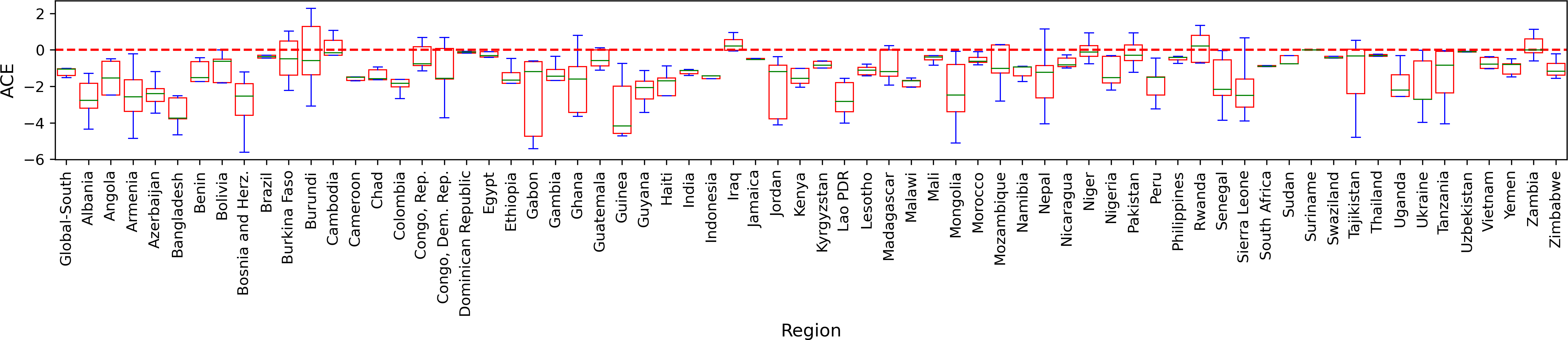}
\caption{
The region-wise average causal effect estimates from 5 random seeded simulations plotted along with the zero-ACE line. 
}\label{sfig:IMF_CP_d32_000_GlobalSouth_deg_sim_d_CACE}
\end{figure*}
In our experiments, we consider the IMF dataset used in~\citet{daoud2019IMFCPeffectheterogeneity, Daoud2017IMFchildhealth, hallerod2013badgovernance_poorchildren} with 1,941,734 children under the age of 18, cared for by 567,344 families residing in 67 countries from the Global-South.
Fig.~\ref{sfig:000_GlobalSouth_TS6_N_samples_cbar} in Appendix~\ref{apdx:IMF_GS_N_samples} shows the country-wise proportions of the observational samples in the IMF dataset.
The benefit of this dataset is that it represents 2.8 billion samples (50\%) of the world's population by the year 2000.
Fig.~\ref{sfig:dagitty_cdag} in Appendix~\ref{apdx:cDAG} shows the c-DAG with 32 variables such as economy, polity, public spending, living condition, etc. along with the observed IMF program treatment at a country level and the degree of the child poverty under the observed treatment considered in our study.
The degree of the child poverty in the dataset is calculated based on 7 individual dimensions of poverty for each of the child based on 
\begin{inparaenum}[1)]
\item education,
\item health, 
\item information,
\item malnutrition,
\item sanitization,
\item shelter,
\item water,
\end{inparaenum}
resulting in an aggregate degree of child poverty between 0 and 7 with 0 indicating no poverty and 7 indicating severe poverty. 
Due to the sensitive nature and the accompanying ethical considerations, the IMF dataset is not made publicly available and can be requested upon from the original authors of~\citet{hallerod2013badgovernance_poorchildren, Daoud2017IMFchildhealth, daoud2019IMFCPeffectheterogeneity}.
Since deep neural networks are prone to overfit, we split the data into 1,922,316 training, 9,709 validation and 9,709 test samples.
We run multiple simulations for the hyperparameter selection and select the best hyperparameters with the best held-out validation loss. 
We strictly use only the training set samples for training and use the held-out validation set for early stopping to get the model with best validation loss. 
We further validate the generalization of the best validation loss model on the held-out test set.
We use three fully-connected layers with [40, 30, 20] hidden units for the graphical conditioner and three fully-connected layers with [15, 10, 5] hidden units for the monotonic UMNN transformer.
We implement c-GNFs in Pytorch~\citep{paszke2017pytorch} using GNF baseline code\footnote{\url{https://github.com/AWehenkel/Graphical-Normalizing-Flows}} and AdamW~\citep{LoshchilovH2019adamw} optimizer with learning-rate=$3e{-4}$ and a batch-size of 1024 (4GB of GPU memory) for all our experiments.

Fig.~\ref{sfig:IMF_CP_d32_000_GlobalSouth_deg_sim_d}(a) shows the child poverty statistics sorted degree-wise for the entire Global-South observed across multiple treatment strategies for multiple simulations. The invariability of the boxplot corresponding to the strategy $TS_{Ob}$ (i.e., naturally observed treatment) validates the consistency assumption of the c-GNF model, as the same outcome is observed for the same observed treatments across multiple simulations. Thus, the c-GNF model satisfies consistency by construction.
More importantly, Fig.~\ref{sfig:IMF_CP_d32_000_GlobalSouth_deg_sim_d}(b) shows the IMF program advisability statistics or the proportions of personalization across multiple treatment strategies. 
Since $TS_{I}$ provides the finest personalization at the individual child level, we observe that IMF program is harmful (discouraged) for 7.25$\pm$4.89\%, beneficial (encouraged) for 61.22$\pm$4.08\%, and neutral for the rest.
Contrast to $TS_{I}$, $TS_{C}$ provides intermediate personalization at country level, hence we observe that IMF program is harmful (discouraged) for 10.8$\pm$7.1\%, beneficial (encouraged) for 88.1$\pm$6.04\%, and neutral for the rest.
It should noted that the invariability of the $TS_{0}, TS_{1}$ and $TS_{C}$ is due to the fact that the treatments for these strategies are fixed beforehand. 

As for $TS_{It}$, our findings also indicate the importance of considering all the seven poverty degrees in the analysis in contrast to~\citet{daoud2019IMFCPeffectheterogeneity} that considers only the indicator of child poverty, as the IMF program may help children move from severe to moderate poverty. This may get obscured if a binary indicator of poverty (poor vs non-poor) is used, leading to the erroneous conclusion that personalization is irrelevant. 
The fundamental oversight from over-simplification by grouping poverty degrees 2 to 7 neglects the improvements within the group, e.g., 7 to 2 is a significant improvement for which the IMF program will be rightly encouraged in $TS_I$. However, $TS_{It}$ considers no change in the indicator of the poverty and hence wrongly assumes that the IMF program is neutral for resource optimization. In other words, $TS_{It}$ wrongly values a change of degree from 2 to 1 more than a change of degree from 7 to 2. 
This is specifically seen in the advisability plots of $TS_{It}$ where, compared to $TS_{I}$, there is an increase in the neutral advisability over the encouragement. 
This important finding of ours reinforces the need of the radical rethinking of~\citet{banerjee2011pooreconomics}'s Nobel Prize winning work stated previously in Section~\ref{sec:intro} that suggests one should not resort to over-simplification of the poor into cartoon characters by ignoring their individual characteristics in all their rich complexities and relative improvements in their living conditions.

Fig.~\ref{sfig:IMF_CP_d32_000_GlobalSouth_deg_sim_d}(c) shows the variation of the average degree of child poverty for all the treatment strategies obtained from averaging Fig.~\ref{sfig:IMF_CP_d32_000_GlobalSouth_deg_sim_d}(a). 
From Fig.~\ref{sfig:IMF_CP_d32_000_GlobalSouth_deg_sim_d}(c), the treatment strategies can be sorted in the decreasing order of their expected degree of child poverty $\mathbf{E}[{Y_{TS_a}]}$: $TS_0 > TS_{It} > TS_{Ob} > TS_1 > TS_C > TS_I$. This indicates that the IMF program is beneficial for the Global-South ($TS_0 > TS_{Ob} > TS_1$).
Fig.~\ref{sfig:IMF_CP_d32_000_GlobalSouth_deg_sim_d}(c) shows that the personalization at the country level due to the treatment strategy $TS_C$ (purple) represents a significant reduction in child poverty over the `one-size-fits-all' treatment strategy $TS_{1}$ (blue) and the sub-optimal naturally observed treatment strategy $TS_{Ob}$ (orange). 
Moreover, personalization at the individual child level is even more beneficial ($TS_C > TS_I$). 

Fig.~\ref{sfig:IMF_CP_d32_000_GlobalSouth_deg_sim_d}(d) reconfirms the beneficial nature of the IMF program across multiple simulations as the average degree of the child poverty is observed to be reduced by $1.2\pm0.24$ degree.
Similarly, Fig.~\ref{sfig:IMF_CP_d32_000_GlobalSouth_deg_sim_d_CACE} shows that most countries experience a reduction in the average degree of child poverty from the IMF program, and hence the program has beneficial effects on the Global-South as a whole.
Figs.~\ref{fig:CACP_ACP_TS_0_1_2_3_4_5} in Appendix~\ref{apdx:worldmap CACP} provides a qualitative visual analysis of the country-wise average degree of child poverty in the form of the six potential (interventional) worlds under each of the six treatment strategies from one of the 5 simulations.
~From these results, we see that it is indeed possible to perform P$^3$A, under the assumptions in Section~\ref{sec:problemdefinition,assumptions,notations}, to effectively combat social ills in contrast to the `one-size-fits-all' approaches. 

\subsection{On interpretability and explanability of the c-GNF}
Since the c-GNF is able to provide the encapsulated-SCM, this enables to go beyond total causal effect estimation and into the mediation analysis dealing with natural/controlled direct/indirect effects~\cite{pearl2009causality}. The probabilistic nature of the c-GNF further enables one to analyse important causal quantities such as probabilities of causation~\cite{pearl1999probabilitiesofcausation} to dive deep into the causal aspects of the social system under study and enable policymakers with personalized policies. Finally, the c-GNF/encapsulated-SCM is able to successfully navigate across all the three rungs of the \emph{`Ladder of Causation'}~\cite{pearl2018bookofwhy}, i.e., (i) association, (ii) intervention, and (iii) counterfactuals, to be able to answer every causal and counterfactual questions using \emph{`The First Law of Causal Inference'}~\cite{pearl2009causality, pearl2018bookofwhy}. Of all the 16 macroeconomic variables that cause child poverty, we observed that political will is the most impactful as it defines the way of the nation and hence impacts in a macroeconomic shock that trickles down to the individual child level, via microeconomic family-level factors. Observing the c-DAG in Fig.~\ref{sfig:dagitty_cdag} of Appendix~\ref{apdx:cDAG}, it is evident that the political will of a country determines other macroeconomic factors such as public spending, polity, governance, and also trickles down to individual child status via intermediate microeconomic factors. For example, consider Pakistan or Iraq where a negative political will is observed and hence the IMF programs are seen to be harmful (recall Fig.~\ref{sfig:IMF_CP_d32_000_GlobalSouth_deg_sim_d_CACE}) due to the tendency of the monetary funds being misused for state-sponsored terrorism instead of the primary objective of economic stability. In contrast, countries such as India and Bangladesh, where positive political will is observed, are seen to be benefiting the most from the IMF program both economy and child poverty-wise. Similarly, for instance, the most number of current CEOs of tech-giants such as Google, Microsoft, Adobe, IBM, Twitter can loosely be attributed to the positive effects of positive political will in India.
It was also observed that the macroeconomic shocks on child poverty can be isolated if the microeconomic factors at the family-level such as family wealth, family head education, number of children/adults in the family are well conditioned. This strengthens the belief of personalizing the treatments/policies to effectively combat/isolate the macro-level shocks trickling down to individual family or child. In summary, our major observations are two-folds: (i) political will is an important macroeconomic factor that defines both macro- and micro- socio-economic status of a nation, (ii) the IMF program coupled with a positive political will is seen to be beneficial for the child poverty status of the country.

\section{Conclusion}\label{sec:conclusion}

In this article, we deployed causal-Graphical Normalizing Flows (c-GNFs) to draw insights from the real-world observational data on the impact of the IMF program on child poverty. 
Our findings in terms of ACE indicated the IMF program to be beneficial for the Global-South, in expectation. 
Apart from considering the traditional `one-size-fits-all' treatment, we also proposed an empirical framework to formulate different personalized treatment strategies at different population granularity levels by computing ACE, country-wise CACE, and child-wise ICE.
Our findings reinforced the radical thinking proposed by~\citet{banerjee2011pooreconomics} to address the problem of child poverty without over-simplification of the poor into cartoon characters.
Even though we demonstrated the personalized treatment strategy formulation framework in a social science setting, our framework can be extended to the field of personalized medicine as well. 
The transparent nature of the c-GNF that is able to provide the encapsulated-SCM enables one to successfully navigate across all the three rungs of the ladder of causation to provide an end-to-end causal inference tool.
\newpage


\bibliography{icml2022}
\bibliographystyle{icml2022}

\appendix
\onecolumn

\section{IMF dataset country-wise observational samples}\label{apdx:IMF_GS_N_samples}
\begin{figure}[ht!]
    \centering
    \includegraphics[width= \textwidth,keepaspectratio]{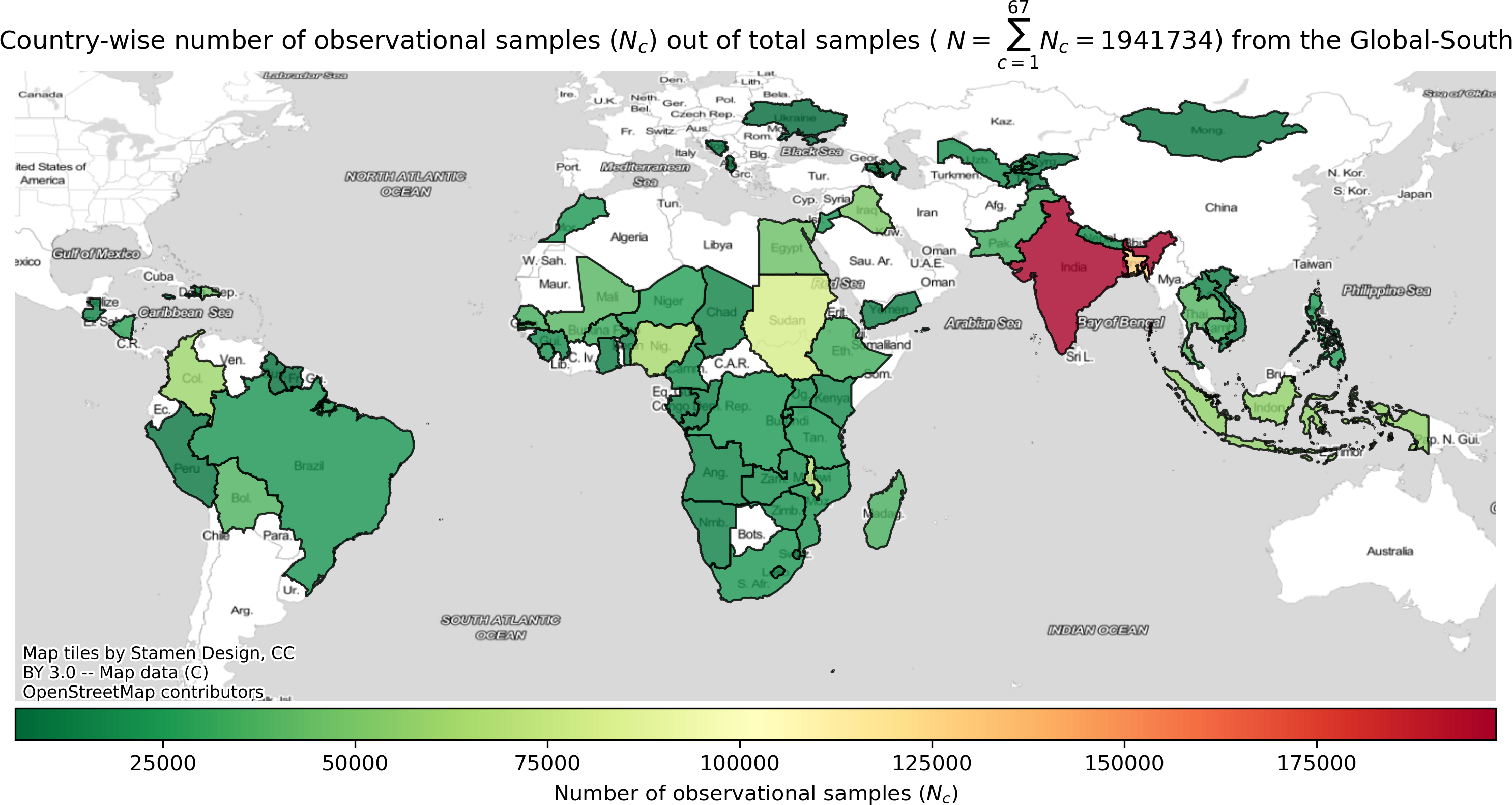}
    \caption{Country-wise proportion of observational samples from all the 67 countries in the Global-South.}\label{sfig:000_GlobalSouth_TS6_N_samples_cbar}
\end{figure}

\section{Summary of the six treatment strategies}\label{apdx: ts_summary}
\begin{table*}[ht!]
\setlength{\tabcolsep}{2pt}
\centering
\begin{tabular}{c|c|c|c|c}
\toprule
\# & Personalization Granularity levels & Treatment assignment condition  & Treatment assigned ($A$) & Treatment strategy  \\ \midrule
1 & None, i.e., entire Global-South level & $0$  & 0 (D) & $TS_0$ (in red) \\
2 & None, i.e., entire Global-South level & $1$ & 1 (E) & $TS_1$ (in blue) \\
3 & Intermediate, i.e., Country level & Naturally observed treatment  & 0 (D) or 1(E) & $TS_{Ob}$ (in orange) \\
4 & Intermediate, i.e., Country level & $\mathbf{CACE}(C{ = }c)$  & 0 (D) or 1 (E) or 0/1 (N) & $TS_C$ (in purple) \\
5 & Finest, i.e., Individual child level & $\mathbf{ICE}_i(C{ = }c, O{ = }o)$  & 0 (D) or 1 (E) or 0/1 (N) & $TS_I$ (in green) \\
6 & Finest, i.e., Individual child level & $\mathbf{ICE}_{it}(C{ = }c, O{ = }o)$ & 0 (D) or 1 (E) or 0/1 (N) & $TS_{It}$ (in olive) \\
\bottomrule
\end{tabular}
\caption{Six IMF treatment strategies considered in our real-world study with the IMF dataset.}
\label{table:imf_ts_6}
\end{table*}

\newpage
\section{Hypothesized causal-DAG for the current social system under study}\label{apdx:cDAG}
\begin{figure*}[hbt!]
\includegraphics[width= \textwidth,keepaspectratio]{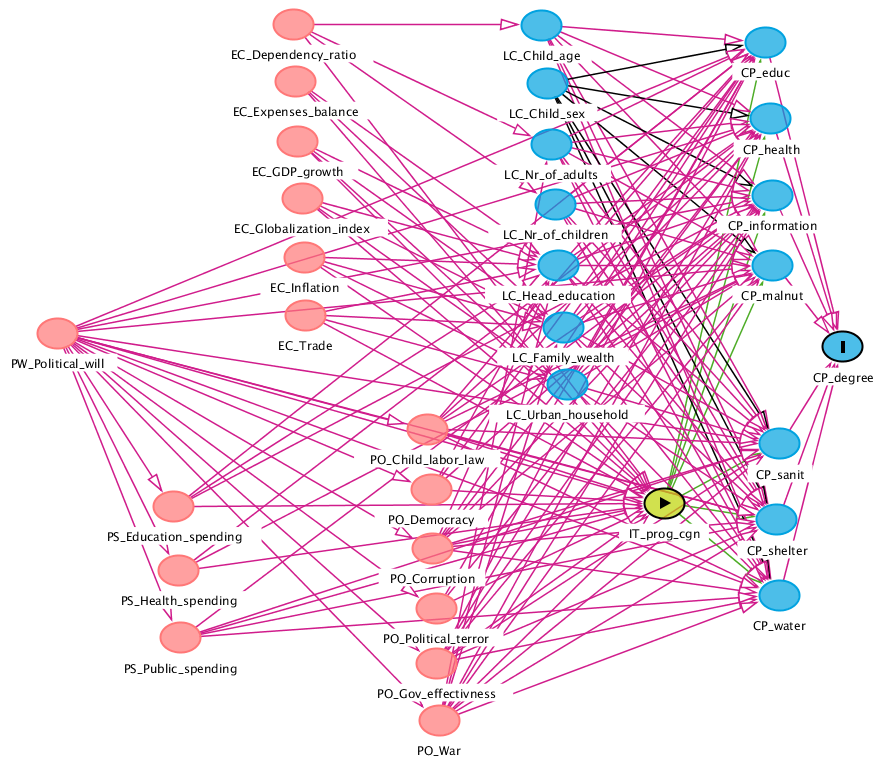}
\caption{
The detailed causal-DAG involving the 32 variables hypothesized for the SCM, created using Dagitty (\url{http://www.dagitty.net/dags.html})~\citep{textor2016dagitty}.
}\label{sfig:dagitty_cdag}
\end{figure*}

\newpage

\section{Counterfactual worldmaps for different treatment strategies}\label{apdx:worldmap CACP}
\begin{figure*}[hbt!]
\subfigure[Counterfactual world under $TS_0$]{\includegraphics[width= 0.49\textwidth,height=\linewidth,keepaspectratio]{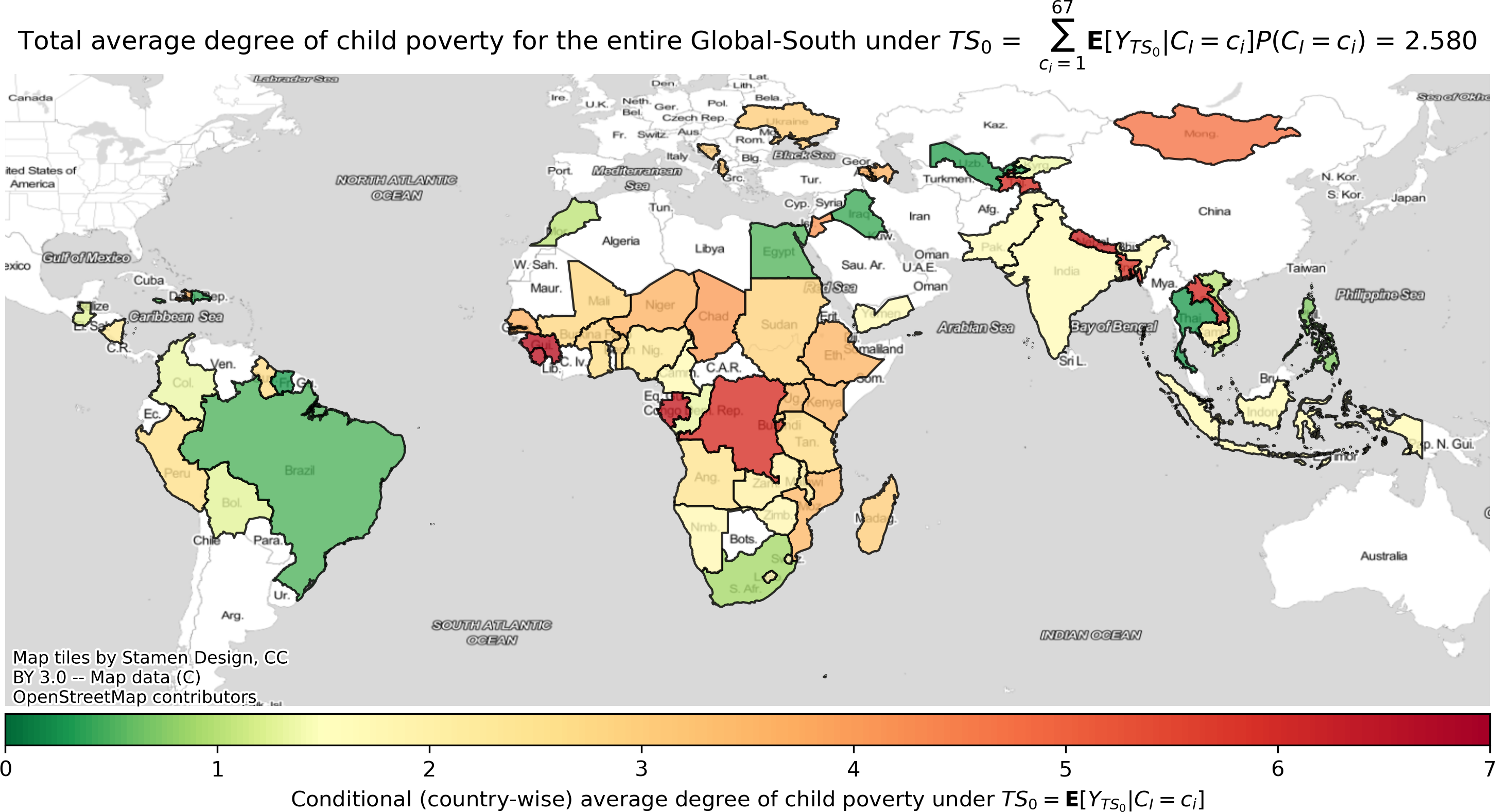}\label{sfig:000_GlobalSouth_CACP_TS6_0_cbar_apdx}}
\hspace{.01\linewidth}
\subfigure[Counterfactual world under $TS_1$]{\includegraphics[width= 0.49\textwidth,height=\linewidth,keepaspectratio]{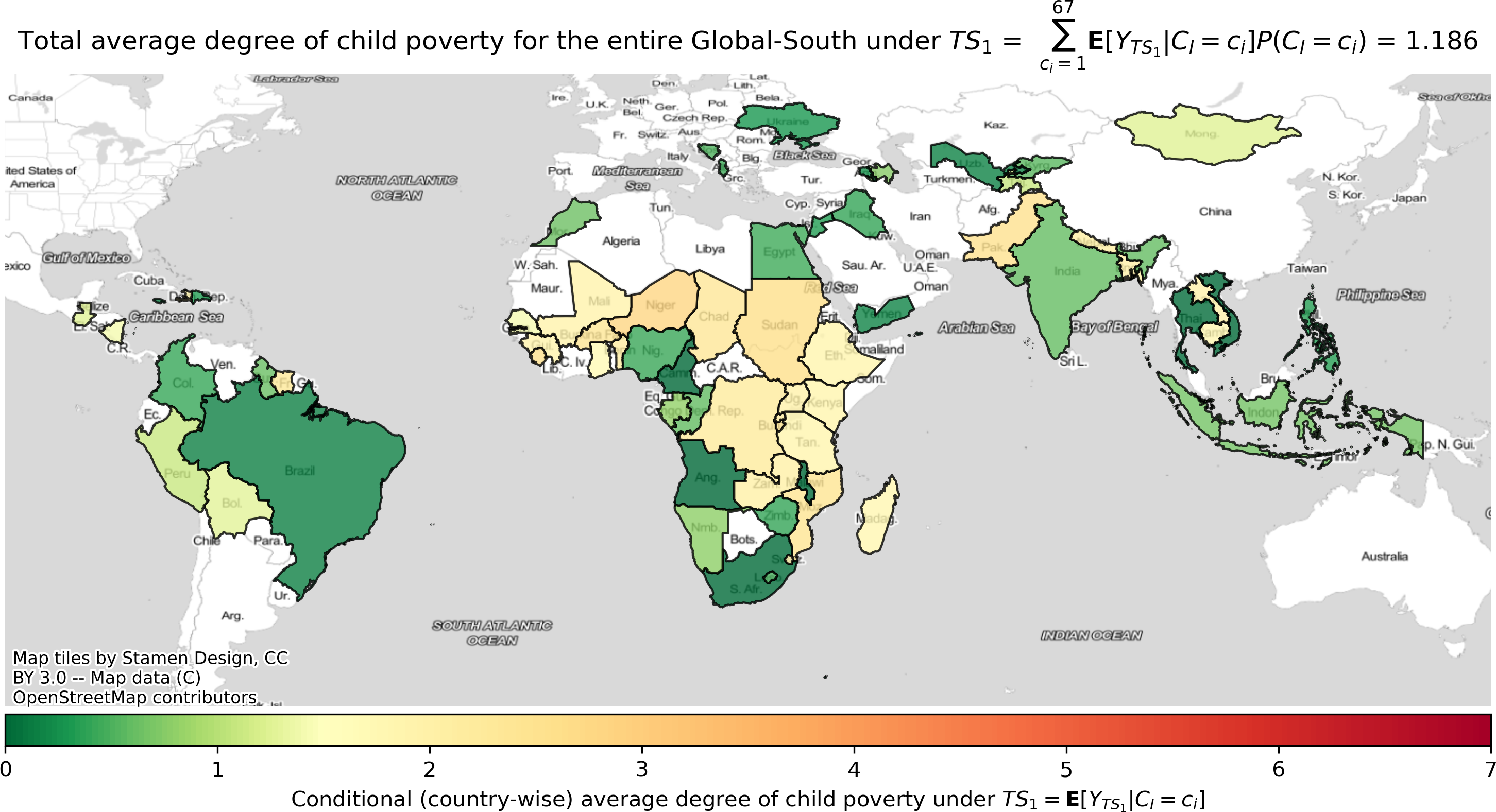}\label{sfig:000_GlobalSouth_CACP_TS6_1_cbar}}
\\
\subfigure[Factual world observed under $TS_{Ob}$]{\includegraphics[width= 0.49\textwidth,height=\linewidth,keepaspectratio]{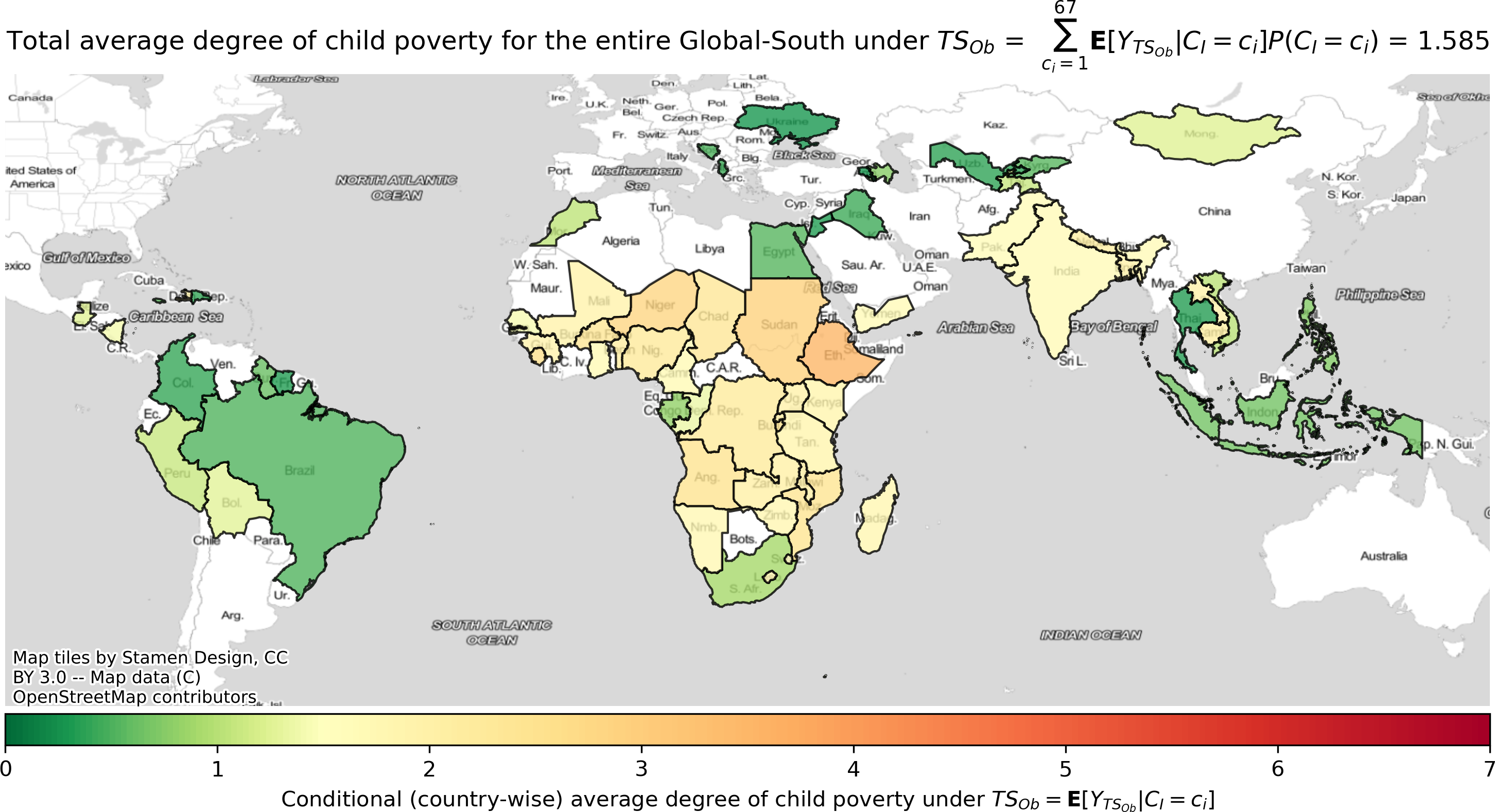}\label{sfig:000_GlobalSouth_CACP_TS6_2_cbar}}
\hspace{.01\linewidth}
\subfigure[Counterfactual world under $TS_C$]{\includegraphics[width= 0.49\textwidth,height=\linewidth,keepaspectratio]{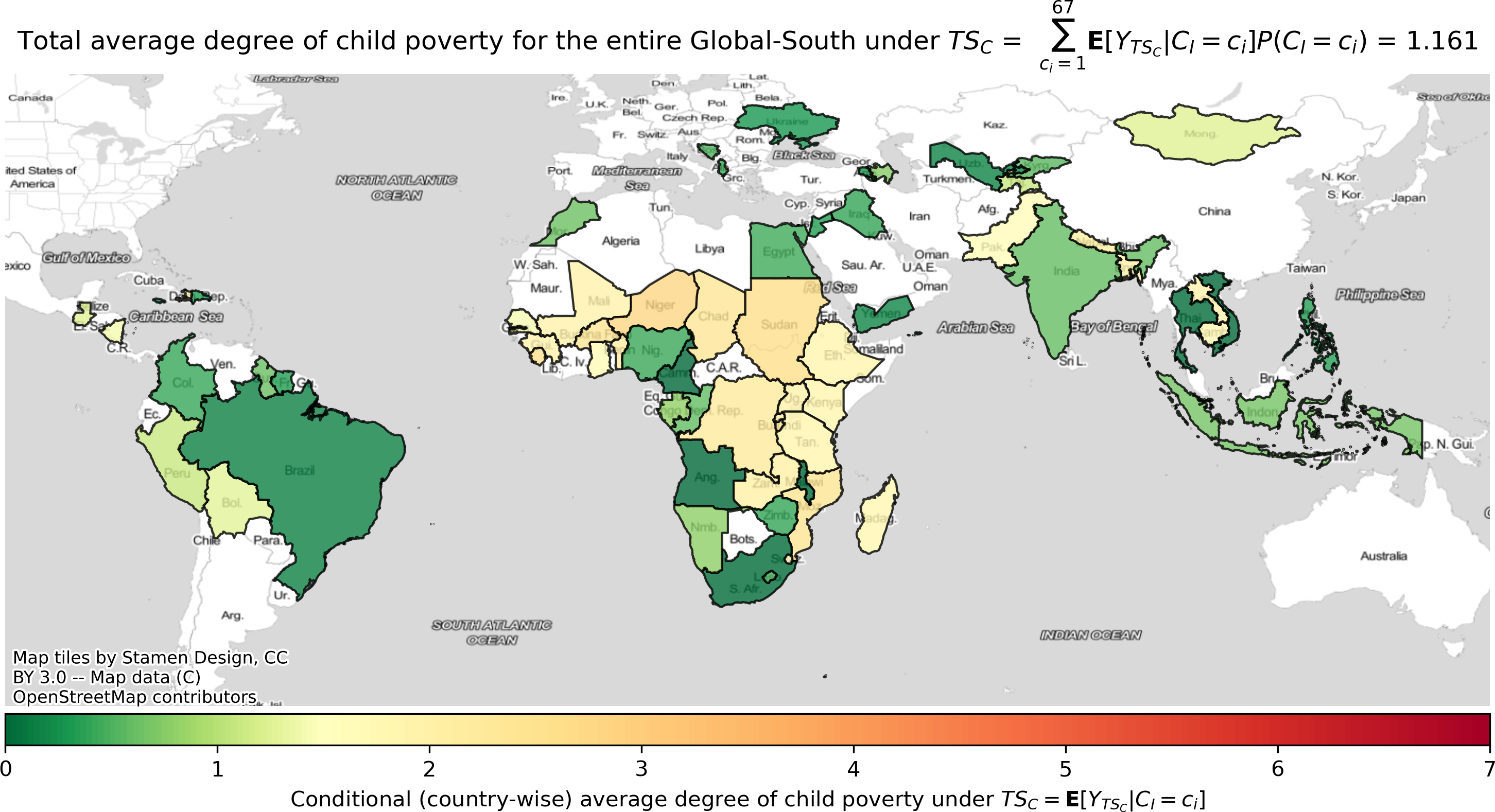}\label{sfig:000_GlobalSouth_CACP_TS6_3_cbar}}
\\
\subfigure[Counterfactual world under $TS_I$]{\includegraphics[width= 0.49\textwidth,height=\linewidth,keepaspectratio]{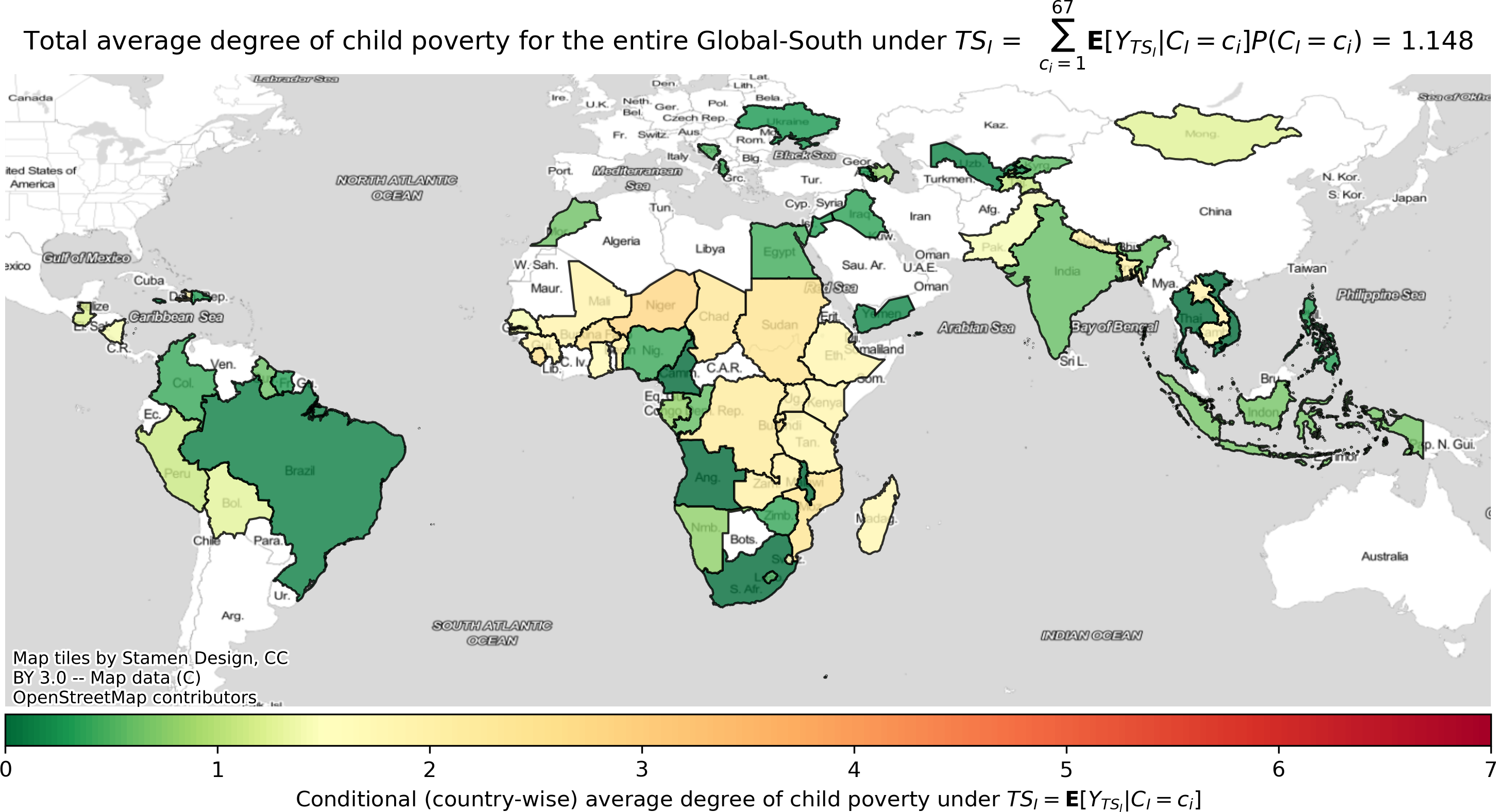}\label{sfig:000_GlobalSouth_CACP_TS6_4_cbar}}
\hspace{.01\linewidth}
\subfigure[Counterfactual world under $TS_{It}$]{\includegraphics[width= 0.49\textwidth,height=\linewidth,keepaspectratio]{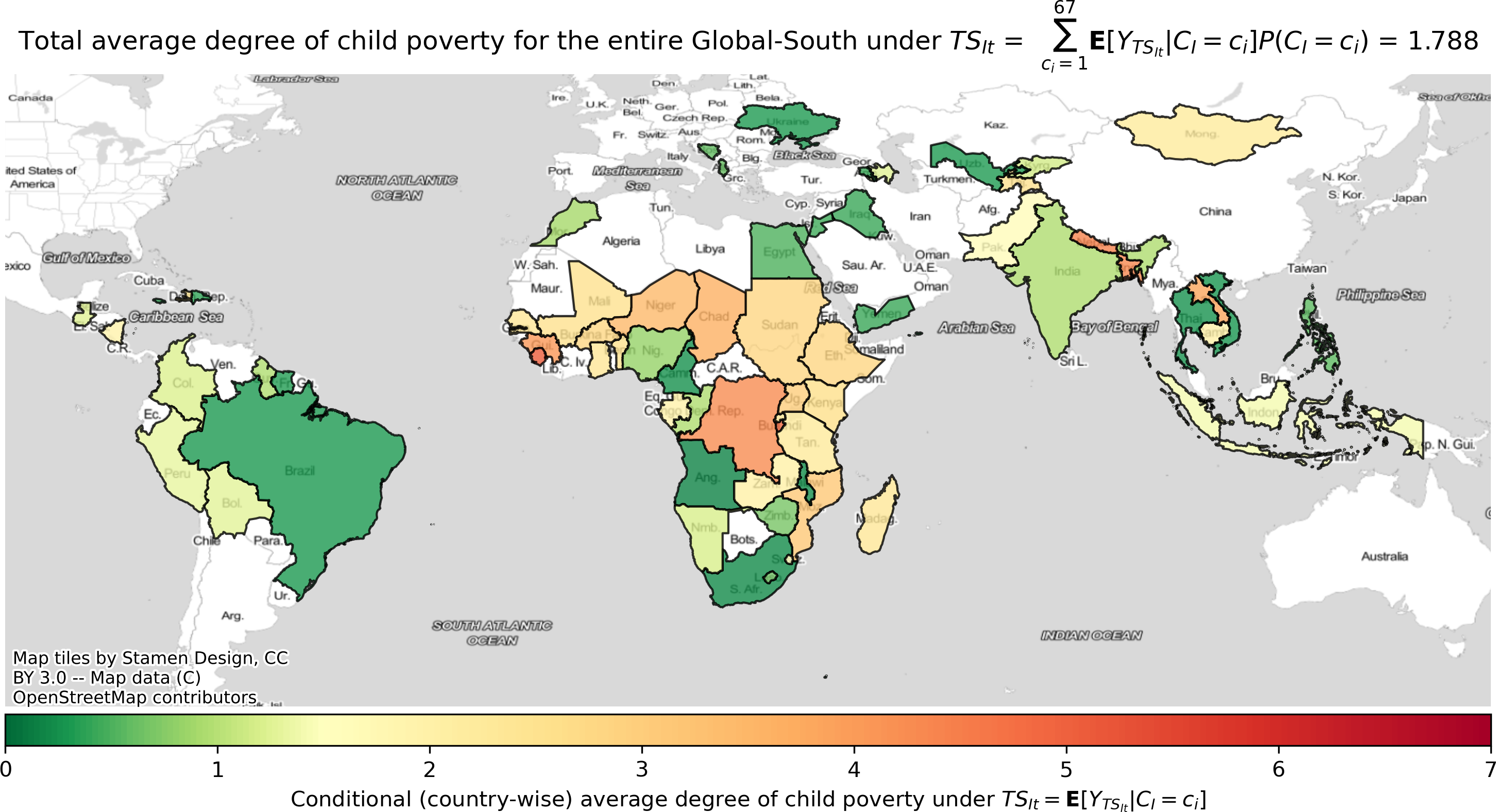}\label{sfig:000_GlobalSouth_CACP_TS6_5_cbar}}
\caption{
Six potential (interventional) worlds under the six treatment strategies indicating the country-wise average degree of child poverty.
}
        \label{fig:CACP_ACP_TS_0_1_2_3_4_5}
\end{figure*}

\newpage

\section{Algorithm for Gaussian dequantization}\label{apdx:GDeq}
\begin{algorithm}[htb!]
   \caption{Gaussian Dequantization}
   \label{alg:dequatization}
\begin{algorithmic}
   \STATE {\bfseries Input:} Discrete variables $\{{D^\ell}\}^b_{\ell{ = }1}$
   \STATE Generate $\tilde{D^\ell} \sim \mathcal{N}(\mu={D}^{\ell},\sigma^2=1/36)$.
   \STATE {\bfseries Output:} Dequantized / continuous variables $\{\tilde{D^\ell}\}^b_{\ell{ = }1}$
\end{algorithmic}
\end{algorithm}
\begin{algorithm}[htb!]
   \caption{Gaussian Quantization}
   \label{alg:quantization}
\begin{algorithmic}
   \STATE {\bfseries Input:} Continuous variables $\{{\tilde{D}^\ell}\}^b_{\ell{ = }1}$
   \STATE Generate ${D}^{\ell} = \mathrm{clamp}(\mathrm{round}(\tilde{D^\ell}), \mathrm{min}{ = }0, \mathrm{max}{ = }N{ - }1)$
   \STATE {\bfseries Output:} Quantized / discrete variables $\{{D^\ell}\}^b_{\ell{ = }1}$
\end{algorithmic}
\end{algorithm}


\end{document}